\documentclass[10pt,twocolumn]{article}
\usepackage{latex8}
\usepackage{times}
\usepackage{examples}
\usepackage{epsfig}

\newenvironment{lquote}{\begin{list}{}{}\item[]}{\end{list}}

\begin{document}

\title{A Real World Implementation of Answer
  Extraction\thanks{This research is funded by the Swiss National
    Science Foundation, project nr. 1214-45448-95}}

\author{Diego Moll\'a Aliod, Jawad Berri, Michael Hess \\
  Computational Linguistics Group \\  
  Department of Computer Science, University of Zurich \\
  Winterthurerstr. 190, CH-8057 Zurich, Switzerland \\
  \{molla,berri,hess\}@ifi.unizh.ch}

\maketitle
\thispagestyle{empty}

\begin{abstract}
  In this paper we describe ExtrAns, an answer extraction
  system. Answer extraction (AE) aims at retrieving those
  exact passages of a document that directly answer a given user question.
  AE is more ambitious than information retrieval and
  information extraction in that the 
  retrieval results are phrases, not entire documents, and in that
  the queries may be arbitrarily specific. It is less ambitious than full-fledged
  question answering in that the answers are not generated from a knowledge
  base but looked up in the text of documents.
  The current version of ExtrAns is able to parse unedited Unix
  ``man pages'', and derive the
  logical form of their sentences. User queries are also translated into
  logical forms. A theorem prover then retrieves the relevant phrases, which
  are presented through selective highlighting in their context.
\end{abstract}

\Section{Answer Extraction: The Core Idea}

One of the fields where natural language understanding technology has failed
to deliver convincing results is \em text based question answering. \em
Systems which read texts, assimilate their content, and answer freely
phrased questions about them would be very useful in a wide variety of
applications, particularly so if questions \em and \em texts could be
written in unrestricted language. They would be the perfect solution to the
problem of information overload in the age of the World Wide Web. However,
as the situation is today (and will remain for a long time to come), such
systems can be implemented only in very small domains, for extremely small
amounts of text, and with very high development costs. One such system,
LILOG~(\cite{Herzog:1991}), absorbed well in excess of 60 person-years of
work and could, in its final stage, still treat merely a few dozen pages of
text. Moreover it turned out to be extremely costly to port the system from
one domain to another, closely related, one (many person-months of work). In
fact these types of systems have become prototypical cases of non-scalable
laboratory applications with very limited impact on further developments.

When it comes to processing larger amounts of texts there have been
around only two serious contenders up until now: Information Retrieval
and Information Extraction. Unfortunately, both techniques have
serious drawbacks. Standard \em information retrieval \em (IR)
techniques allow arbitrary queries over very large document
collections (many gigabytes in size) covering arbitrary domains but
they usually retrieve \em entire documents \em (this holds true for
traditional systems such as SMART~\cite{Salton:1989} as well as for
probabilistic ones such as SPIDER~\cite{Schaeuble:1993}). However,
this is unhelpful if documents are dozens, or hundreds, of pages long.
Sometimes the techniques of IR are also used to retrieve individual
passages of documents (one or more sentences, paragraphs)
(cf.~\cite{Salton:1993}). In such cases the number of search terms
found in a given sentence, together with their density (in terms of
closeness in a sentence), is used to find relevant sentences.

Unfortunately, all IR techniques (whether applied to entire documents or to
individual passages) have a number of limitations that make them unsuitable
for certain important applications. First, they take into account only the
content words of a document (all the function words are thrown away).
Second, in most cases only the stem of such words is used (and this stem is
usually not derived by a proper morphological analysis but by means of some
kind of stemmer algorithm, inevitably resulting in numerous spurious
ambiguities). Finally, and most importantly, the resulting terms are treated
as \em isolated items \em whose unordered combination is used as content
model of the original document. This holds for Boolean systems as well as
for vector space based systems. Inevitably, neither model can, as such,
distinguish the concept of ``computer design'' from that of ``design
computer'' (lost ordering information), or the concept of ``export from
Germany to the UK'' from that of ``export from the UK to Germany'' (lost
function word information). True, most systems can use phrasal search terms
(such as "computer design"), to be found as a whole in the documents, but
then a number of relevant documents (such as those containing "design of
computers") will no longer be retrieved. All this also holds for the (few)
passage retrieval systems described in the literature (such as the system
described in~\cite{Salton:1993}). 

\em Information extraction \em (IE) techniques do not suffer from the same
shortcomings. They are similar to IR systems in that they, too, are suitable
for screening very large text collections (of basically unlimited size, such
as streams of messages) covering a potentially wide range of topics.
However, they differ from IR systems in that they not only identify certain
messages in such a stream (those that fall into a number of specific topics)
but also extract from those messages highly specific content data. Typical
examples are newswire reports describing terrorist attacks (where they
extract the information as to who attacked whom and how and when, what was
the outcome of the attack etc.) or newswire reports on management
succession events in newswire business reports (with data on who resigned
from what post in which company, who is successor etc.). This predefined
information is placed into a template, or data base record, defined for the
different role fillers of a given type of report.

Clearly, this kind of information is much more precise and specific than
what is considered by IR systems. On the other hand, IE systems do not allow
for arbitrary questions (as IR systems do). They merely allow for a small
number of \em pre-defined \em information frames to be filled. Worse still,
the ``Message Understanding Conferences'', which have been driving
development in this area since 1987 (the latest so far, with published
proceedings, is MUC-6~\cite{ARPA:1996}), put so much emphasis on very large
text volumes that most of the participating systems that had used, at
first, a thorough linguistic analysis had to abandon it and adopt a
very shallow approach instead, simply because of run-time requirements for
such volumes of data~(e.g. \cite{Appelt:1993}). This approach, which is now taken
by most systems taking part in MUCs, makes the systems increasingly less
general.

However, there is a growing need today for systems that are capable of
locating information in texts \em not \em running into the gigabytes but
which should show very high precision and recall and which should
furthermore allow \em arbitrarily phrased \em questions. Moreover they
should be able to cope with documents written in syntactically \em
unrestricted \em natural language whereas the \em domain \em of the texts is
normally quite \em restricted \em. Examples for such systems are interfaces
to machine-readable technical manuals, on-line help systems for complex
software, help desk systems in large organisations, and public inquiry
systems accessible over the Internet. For these tasks, very high precision
of retrieval is mandatory (queries may be very specific), often near perfect
recall is vital (technical manuals typically explain things only once), and
sometimes retrieval time is mission critical (retrieving information about a
system about to get out of control). What is needed in such situations is a
system that pinpoints the exact \em phrase(s) \em in a document (collection)
from whose \em meaning \em we can infer the answer to a specific question.
This is the core idea of Answer Extraction (AE). Since we need to determine
the meaning of sentences (questions and texts) we must use (a limited degree
of) linguistic (syntactic and semantic) information, which is expensive, but
on the other hand the texts to be processed are moderately sized (some
hundreds of kilobytes, sometimes a few megabytes), and they typically cover
a very limited domain. This makes Answer Extraction a realistic compromise
between full question answering on the one hand, and mere information
extraction or information retrieval on the other.

We will describe an Answer Extraction system, ``ExtrAns'', for questions
about (a subset of) the on-line Unix manual (the so-called ``man pages'').
Although the system is, for the time being, functional only as a prototype
it can cope with unedited text and arbitrary questions, with performance
degrading gracefully if input (documents or questions) cannot be analysed
completely. It is incrementally extensible in the sense that refinements of
the grammar and/or the semantic component automatically improve precision
and recall, without the need to change any other components of the system.

\Section{Requirements and components} 
\label{sec:arch}

Given the fact that an Answer Extraction system should be able to cope with
unrestricted text, it needs a very reliable tokeniser, a grammar of
considerable coverage, a reasonably robust parser, some way of dealing with
ambiguities, a module that can subject even fragments of syntax structures
to a semantic analysis, and a search engine capable of using the resulting
knowledge base.

In the following we will describe merely three components of the
system in some detail. \em First, \em we will point out that
preprocessing technical language goes well beyond what a typical
tokeniser does. We will not describe the syntax analysis module for
which we use and extend an existing dependency oriented system that
comes with a full form lexicon, a grammar, and a parser, viz.  Sleator
and Temperley's ``Link Grammar''~\cite{Sleator:1991}.  It has certain
built-in capabilities for robust parsing, which we supplement by a
fall-back strategy that turns unrecognised constituents into keywords
(thus resorting to an IR-type behaviour). \em Second, \em we will
describe the design principles for the semantic representations
derived from the (very specific type of) syntax structures produced by
Link Grammar.  We will not explain in depth the disambiguation module,
for which we adopt and extend the approach put forward by Brill
and Resnik~\cite{Brill:1994}.  \em Third, \em we will show how we cope
with the syntactic ambiguities that survive all our disambiguation
activities. We will also explain the search strategy very briefly.

\Section{Preprocessing technical language} 
\label{sec:tokeniser}

The analysis of technical language is, in general, considerably simpler than
that of domain unspecific language (newspapers etc.) but as far as
preprocessing is concerned it is far more demanding. This holds, in
particular, for tokenisation, normalisation, and document structure
analysis.

\SubSection{Tokenisation and normalisation}

In general, tokenising a text means merely identifying word forms and
sentences. However, in highly technical documents such as the Unix man
pages, this may become a formidable task. Apart from regular word forms, the
ExtrAns tokeniser has to recognise all of the following as tokens and
represent them as normalised expressions:

{\raggedright
\paragraph{Command names:} eject, nice (problem: identify regular words
    when used as names of commands in sentences like ``\bfseries eject
    \mdseries is used for...'', as opposed to their standard use, as
    in ``It is not recommended to physically eject media ...'').

\paragraph{Path names and absolute file names:} /usr/bin/X11; usr/5bin/ls,
    /etc/hostname.le (problems: leading, trailing and internal
    slashes, numbers and periods).

\paragraph{Options of commands:} -C, -ww, -dFinUv (problem: identify where
    a sequence preceded by a dash is an option and where not, as in
    ``... whose name ends with .gz, -gz, .z, -z, \_z or .Z and which
    begins ...'').

\paragraph{Named variables:} \em filename1, device, nickname \em (problems:
    identify words used as named variables, mostly as arguments of
    commands as in ``... the first \em mm \em is the hour number; \em dd
    \em is the day ...'')

\paragraph{Special characters as parts of tokens:} AF\_UNIX, sun\_path,
    \^{}S(CTRL-S), KR, C++, name@domain or 
    \%, \%\% (as in: ``A single \%
    is encoded by \%\%.''), various punctuation marks (as in: ``...
    corresponding to cat? or fmt?'', or in ``/usr/man/man?'', 
    ``\(<\)signal.h\(>\)'', or 
    ``[host!...host!]host!username '')

}

\vspace{1ex}

Normalising such tokens means, among other things, to appropriately mark
special tokens such as command names (otherwise the parser chokes on them).
Luckily, the Unix man pages contain a considerable amount of useful
information beyond the purely textual level, namely the information conveyed
by the \em formatting commands \em. Thus command names are, as a rule,
printed in boldface, and expressions used as variables, in italics, as in
\begin{lquote}
  \textbf{compress} [ -cfv ] [ -b \em bits \em ] [ \em filename \em...]
\end{lquote}

This type of information is extracted from the formatting commands and added
to the tokens for later modules to use (e.g. ``eject'', when used as the
name of a command, is turned into ``eject.com'', and ``filename'', when used
as an argument, into ``filename.arg'').

\SubSection{Document structure analysis}

The formatting instructions in the Unix man pages are, unfortunately,
used in a fairly unsystematic fashion (these pages were written by
dozens of different persons). In order to extract additional
information about tokens from the formatting (see above), the
tokeniser must make up for these inconsistencies in the source texts.
It does so by performing a considerable amount of document structure
analysis. It has, for instance, to collect all command names and
argument names from the SYNOPSIS and NAME sections of each manual page
to be sure that it will recognise all of them in the body of the
DESCRIPTION section, even if formatted incorrectly. Thus, processing a
man page becomes a case of processing each of its sections in a
particular way.

\Section{Logical forms} 
\label{sec:logform}

A major property of ExtrAns is that the textual information is converted
into \textbf{existentially closed formulae of logic}, which encode the main
content relationships between the words of the sentences. All formulae are
existentially quantified since, for retrieval purposes, all entities
mentioned in a document can be assumed to exist, as generic entities, in the
universe of discourse shared by writer and reader. Verbs, nouns, adjectives,
and adverbs thus introduce entities in the universe of discourse which can
be referenced later. In particular, the verb ``copy'' in ``cp copies good
files'' introduces, in the predicate \texttt{evt(copy,e1,[c1,f1])}, an entity
\texttt{e1} representing the concept of copying. (\texttt{e1} can be seen as
a reified eventuality in the sense of~\cite{Parsons:1990}). We apply this
notion of reification to other predicates also. The noun ``cp''
introduces a predicate \texttt{object(cp,o1,c1)}, where \texttt{c1}
represents the command \textbf{cp} itself while \texttt{o1} stands for the
\em concept \em of \texttt{c1} being of the type \textbf{cp}. Similarly, the
adjective ``good'' introduces, in \texttt{prop(good,p1,f1)}, a new concept,
\texttt{p1}, viz. the  \em concept  \em of \texttt{f1} having the
property \textbf{good}. These additional entities (\texttt{o1,p1}) can be
used to model intensional constructions like ``X is an alleged copy'',
where the concept of X's being a copy is qualified (as opposed to,
say, saying that X is a copy and X is also alleged), or ``Y is pale
green'', where Y's being green is modified.

As a basic knowledge representation language we use \textbf{Horn
  Clause Logic}, the subset of Predicate Logic which can be directly
handled by Prolog. We adopt a mixed-level ontology (largely following
\cite{Hess:1997}): For each main verb we create one fixed-arity
predicate while its (obligatory) complements and (non-obligatory)
modifiers result in additional predicates. The resulting expressions
are supplied with pointers to the sentences and individual word forms
from which they were derived.

The following two examples illustrate an additional distinction we make: 

\begin{examples} 
\item ``\textbf{cp} copies the contents of \em filename1 \em onto \em filename2\em''
\label{ex:copy}
\item ``If the operation fails, \textbf{eject} prints a message''
\label{ex:pr} 
\end{examples} 

In the first example the copying event is asserted to actually hold (in the
generic world of man pages), but this is not the case for either of the
actions in the \em second \em example (failing and printing) since both of
them are introduced in the scope of the conditional. Clearly we want
passage~(\ref{ex:pr}) to be retrievable for appropriate queries (see example
below) but presumably we do \em not \em want it to be shown when we ask
about ways for the \em user \em to \em intentionally \em print a message, as
opposed to the printing of error messages, which is merely a side-effect.
For this reason we further extend the ontological basis of our system concerning
eventualities and state that those eventualities that actually hold in the
world are explicitly marked as such. All others are presumed to merely exist
in the universe of discourse.\footnote{Note that we do not do this yet for
objects and properties, though.} We get thus, for~(\ref{ex:copy}), the following
representation\footnote{The pointers to word positions are not shown here.}
\begin{lquote}
  {\small \texttt{holds(e1)/s1. object(cp,o1,x1)/s1. \\
      object(command,o2,x1)/s1. \\
      evt(copy,e1,[x1,x2])/s1. \\
      object(content,o3,x2)/s1. \\
      object(filename1,o4,x3)/s1.  \\
      object(file,o5,x3)/s1. of(x2,x3)/s1. \\
      object(filename2,o6,x4)/s1. \\
      object(file,o7,x4)/s1. onto(e1,x4)/s1.}  } 
\end{lquote}
where the copying event is said to hold
(\texttt{holds(e1)}), 
and, for~(\ref{ex:pr})
\begin{lquote} {\small
    \texttt{object(eject,o8,x7)/s2. \\
      object(command,o9,x7)/s2. \\
      evt(print,e8,[x7,x11])/s2. \\
      object(message,o10,x11)/s2. if(e8,e5)/s2.  \\
      object(operation,o11,x5)/s2. \\
      evt(fail,e5,[x5])/s2.} } 
\end{lquote} 
where neither
the failing nor the printing event (\texttt{e5} and \texttt{e8}) is marked
that way. Actual existence may be inferred or blocked or let unspecified,
according to context \cite{Hobbs:1985}.

As we can see above, the logical forms generated are simplified. The
conditional, for example, is not encoded as logical implication but as a
regular predicate \texttt{if( , )}. The same holds for negation, which is
introduced as another regular predicate \texttt{not( )}. Further
simplifications are that plurals, modality, tense, and quantification are
ignored.  As a result, there is an obvious decrease in precision, but the
retrieval results are perfectly sensible as a rule. In fact, in some cases,
even if a retrieved sentence does not strictly answer the answer to the
query, it does provide useful information to the user. For example, if the
user asks ``Which commands can print warning messages?'' (note the plural and
the modal), the system can easily retrieve~(\ref{ex:pr}), which contains a
conditional whose antecedent cannot be proven to hold, but still the
sentence by itself is useful as an answer. Had we used a more detailed
logical form, the system would have to resort to possibly complex inferences
in order to retrieve the same result. Research in this area is not yet
finished, however, and it is possible that more complex logical forms are
used in further prototypes of ExtrAns.

Finally, note also that we have used in this representation some of
the information extracted by the tokeniser from the \em typography \em
(and, indirectly, from the document structure) of the text. The fact
that ``eject'' is the name of a command (rather than a regular English
word) results in the creation of an additional entry
\texttt{object(command,x7)}, and similarly ``cp'' is recognised as
referring to a command, too. Without recourse to this type of
information the first sentence would become unavailable for queries
like ``What commands copy files?'', and the second sentence would have
to be considered ungrammatical.

In order to answer questions 
the standard search procedure of refutation resolution (as
implemented in Prolog) is now used to find \em all
\em proofs of the query over the knowledge base. The query 
\begin{examples}
  \item Which command copies files?
\end{examples}
thus becomes, after computing the logical form and checking for
synonyms (see next section)
\begin{lquote}
  \begin{tabbing}
\small\texttt{?- findall(S,(}\=\small\texttt{object(command,$\_$,X)/S,}\\
 \>\small\texttt{(}\=\small\texttt{evt(copy,E,[X,Y])/S;}\\
 \>                \>\small\texttt{evt(duplicate,E,[X,Y])/S;}\\
 \>\small\texttt{object(file,$\_$,Y)/S), R).}
\end{tabbing}
\end{lquote}
and returns, in \texttt{R=[S1,S2,...]}, the references to the
relevant sentences, one of which is~(\ref{ex:copy}).

\Section{A fall-back search strategy}

Another important feature of ExtrAns is the linguistically aware
fall-back search strategy it uses. For that effect, we are developing
a custom-made, WordNet-style thesaurus~\cite{Miller:1990} which
contains two types of relations between the concepts in the world of
Unix man pages: synonyms and hyponyms.

The overall search algorithm is as follows.  First, all the \em
synonyms \em of all the content words used in the query are added to
the query from the start. If the query does not return enough answers,
\emph{hyponyms} of all search terms will be used. If this, too, gives
us an insufficient number of hits, all the \emph{logical dependencies
  between terms} are broken. At this stage, a query like ``How can I
create a directory?''  would retrieve ``mkdir creates a new directory
file''. Finally, if everything else fails, we go into \emph{keywords}
mode. In this mode, nouns, verbs, adjectives and adverbs are selected
in both the query and the data. Then, every word selected in the
query is matched against those in the document. The result is
displayed according to the number of words in the sentence that match it.
Note that this keywords mode is still more powerful than the standard
IR approach since (\emph{i}) we make good use of the part-of-speech of
the words to decide what is elligible as a keyword, (\emph{ii}) we
make use of the information obtained by the tokeniser from the
formatting of the document (such as names of commands, types of
arguments, etc.), and (\emph{iii}) we also make use of the textual
structure of the document (all index terms have to occur in the same
sentence in order to be considered).

\Section{Presenting (possibly ambiguous) results} 
\label{sec:answer}

A problem that our system (as every NLP system) had to confront was
that of ambiguities. Sleator and Temperley's~\cite{Sleator:1991}
parser does not try to resolve syntactic or semantic ambiguities, and
a long sentence may have hundreds, or even thousands, of different
parses. ExtrAns tries to resolve (syntactic) ambiguities in two steps.
First of all, some hand-crafted rules filter out the most
straightforward cases of spurious ambiguities. An example of such a
rule is: ``A prepositional phrase headed by \emph{of} can attach only
to the immediately preceding noun or noun coordination.'' In a second
step, we adopt Brill and Resnik's~\cite{Brill:1994} prepositional
phrase disambiguation approach, trained with data extracted from
manual pages. With the help of this disambiguator we can use
statistical data to resolve some of the prepositional phrase
attachment ambiguities, a major source of structural ambiguity.

\begin{figure*}
  \begin{center}
    \leavevmode
    \epsfig{file=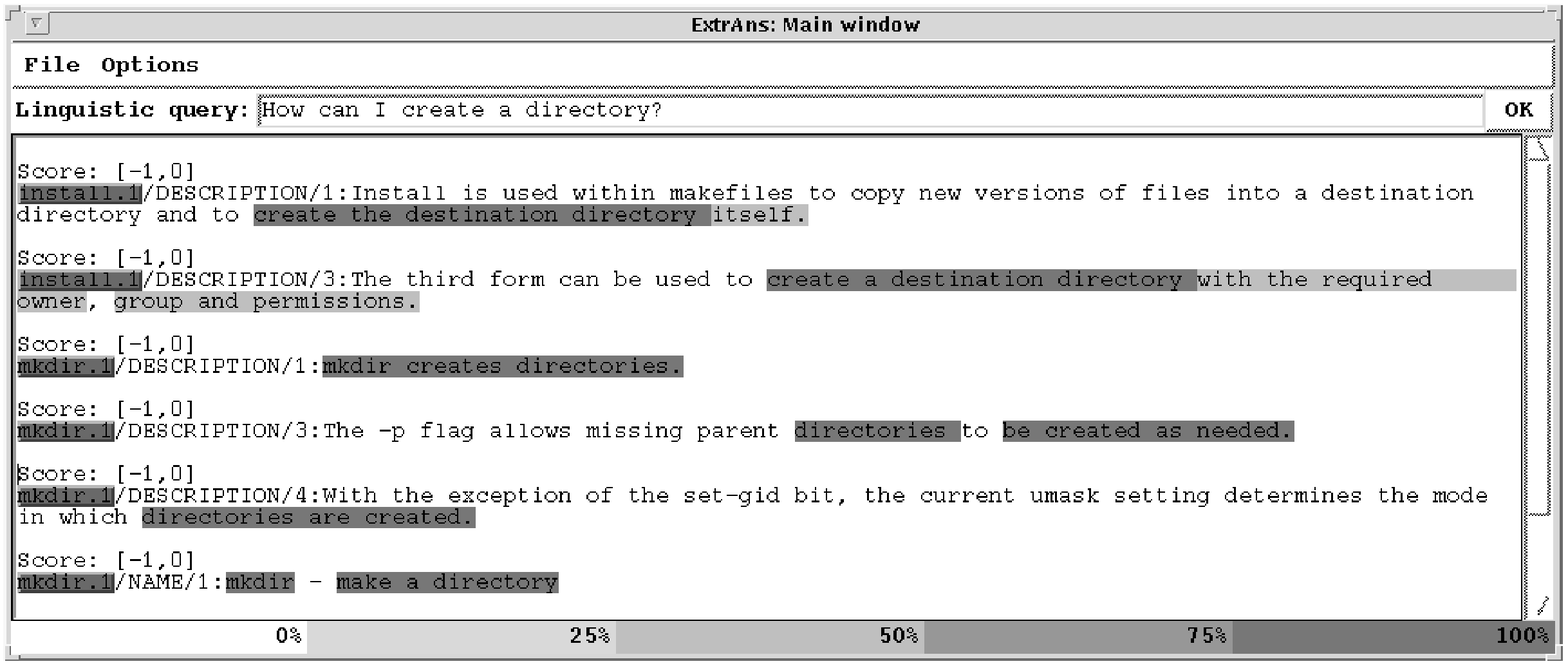,width=\textwidth}
    \caption{The result of the query ``How can I create a directory?'' 
      (original in colour)}
    \label{fig:demo1}
  \end{center}
\end{figure*}

After these two steps, the number of ambiguities will be reduced but some
will normally survive, partly because there are sources of ambiguities which
cannot be treated with Brill and Resnik's algorithm. If a sentence has
several irreducible interpretations, ExtrAns stores \em all \em of them in
its database. When the user asks a query, the same sentence may therefore be
retrieved several times, via different proof paths. The different proofs may
result in the highlighting of different words in the same sentence. ExtrAns
handles this mismatch of highlighted words by superimposing all of the
highlights of a given sentence, using a \emph{graded colouring scheme} in
such a way that those parts which are retrieved several times are
highlighted with a brighter colour than others. For example, consider the
string ``create the destination directory'' in the first hit in
figure~\ref{fig:demo1} (``install.1/DESCRIPTION/1'').
This string is part of \em all \em interpretations of the (ambiguous)
sentence in ``install.1/DESCRIPTION/1'' and was thus used for \em all \em
the proofs of the user query. As a consequence, it is highlighted with highest
intensity in the answer window. 
This means that, in formal terms, we interpret unambiguity as
relevance. Unconventional as this may be, it seems a very helpful concept in
the face of unreducible ambiguities.

In addition, it is possible to access the complete manual page
containing the sentence by clicking on the manual page name at the
left of each sentence.  The manual page will show the same
multi-coloured selective highlighting, thus enabling the user to spot
the relevant sentence at once and to determine even better, by
inspecting the context, whether the sentence contains in fact an
answer to the question. This way of presenting search results makes
even multiple ambiguities fairly unobtrusive.

\Section{Conclusions and further research} 

\paragraph{Comparison with existing approaches.}

We have built a small prototype that currently processes 30 UNIX manual
pages and allows the user to ask questions in plain English. Since the
amount of data is still small, a statistically \em meaningful \em evaluation
is out of the question. Moreover, it is unclear how we should compare the
performance of our system with that of standard IR systems. The standard
measures of recall and precision for those systems are based on experts'
judgement concerning the relevance of entire documents whereas, in ExtrAns,
we would have to determine the relevance of individual phrases. However, in
an informal manner we can compare our approach and the standard IR approach
by telling ExtrAns to go into keywords mode from the very
beginning. It would then
regularly find a considerable number of passages which are far from
relevant to the query. For example, the query used in figure 1, ``How
can I create a directory?'', would now find (in addition to all
the relevant passages) sentences such as ``\textbf{ln} creates an additional
directory entry, called a link, to a file or directory.'' (which is \em not
\em about the creation of directories) as well as the equally irrelevant ``A
hard link is a standard directory entry just like the one made when the file
was created.'' Ignoring the syntactic, and hence semantic, relationships
between the individual words resulted, predictably, in a considerable loss
in precision. Since even our keywords mode is far more restrictive than
the standard IR search model (see above), any standard IR system is bound
to show considerably lower precision than our AE approach. 

In sum, we hope to have shown that 
the \emph{concept} of answer extraction is very useful, and that it requires
a relatively \emph{limited amount of language processing.} We think we could also
show that it is surprisingly easy for users to cope with ambiguities in
documents if
they are \emph{fused, graded, and presented in context}.

\vspace{-0.2ex}
\paragraph{Features to improve.}

For the system to be truly useful we must clearly increase the number
of manual pages which can be analysed.  We must then refine the
system's treatment of ungrammatical text. Currently, it converts words
which cannot be analysed into isolated keywords, and we should refine
this method so that we can process individual \emph{phrases} if a
sentence cannot be parsed in its entirety. Also, in some sentences,
the number of unreducible syntactic analyses still is overwhelming,
and it will be necessary to refine our methods to disambiguate and
filter out the implausible meanings, possibly through the use of
semantic information.  We must also add some capability of resolving
pronouns and anaphoric full noun phrases. Finally, we must extend the
current inferencing techniques to increase recall. We currently
integrate synonymy, hyponymy and conjunction distributivity, among
others, but we still need to add more inferences and extend the
thesaurus.


\end{document}